\PassOptionsToPackage{table}{xcolor}
\documentclass[sigconf]{acmart}
\AtBeginDocument{%
  }

\setcopyright{cc}
\setcctype{by}
\copyrightyear{2026}
\acmYear{2026}
\acmDOI{10.1145/3767308.3835848}
\acmConference[MM '26]{Proceedings of the 34th ACM International Conference on Multimedia}{November 10--14, 2026}{Rio de Janeiro, Brazil}
\acmBooktitle{Proceedings of the 34th ACM International Conference on Multimedia (MM '26), November 10--14, 2026, Rio de Janeiro, Brazil}
\acmISBN{979-8-4007-2213-4/2026/11}

\usepackage{enumitem}
\usepackage{multirow}
\usepackage{colortbl}

\title[AffectAgent: Multi-Agent Reasoning for Multimodal Emotion Recognition]{AffectAgent: Collaborative Multi-Agent Reasoning for Retrieval-Augmented Multimodal Emotion Recognition}

\author{Zeheng Wang}
\affiliation{%
  \institution{Guangdong Laboratory of Artificial Intelligence and Digital Economy (SZ)}
  \city{Shenzhen}
  \state{Guangdong}
  \country{China}
}
\affiliation{%
	\institution{Great Bay University}
	\city{Dongguan}
	\state{Guangdong}
	\country{China}
}

\author{Zitong Yu}
\authornote{Corresponding author.}
\affiliation{%
  \department{}
  \institution{Great Bay University}
  \city{Dongguan}
  \state{Guangdong}
  \country{China}
}
\affiliation{%
	\department{Guangdong Provincial Key Laboratory of Intelligent Information Processing \& Shenzhen Key Laboratory of Media Security}
	\institution{Shenzhen University}
	\city{Shenzhen}
	\state{Guangdong}
	\country{China}
}

\author{Yijie Zhu}
\affiliation{%
  \institution{Great Bay University}
  \city{Dongguan}
  \state{Guangdong}
  \country{China}
}

\author{Bo Zhao}
\affiliation{%
  \institution{Great Bay University}
  \city{Dongguan}
  \state{Guangdong}
  \country{China}
}

\author{Haochen Liang}
\affiliation{%
  \institution{Great Bay University}
  \city{Dongguan}
  \state{Guangdong}
  \country{China}
}

\author{Taorui Wang}
\affiliation{%
  \institution{Great Bay University}
  \city{Dongguan}
  \state{Guangdong}
  \country{China}
}

\author{Wei Xia}
\affiliation{%
  \institution{Great Bay University}
  \city{Dongguan}
  \state{Guangdong}
  \country{China}
}

\author{Jiayu Zhang}
\affiliation{%
  \institution{Great Bay University}
  \city{Dongguan}
  \state{Guangdong}
  \country{China}
}

\author{Zhishu Liu}
\affiliation{%
  \institution{Great Bay University}
  \city{Dongguan}
  \state{Guangdong}
  \country{China}
}

\author{Hui Ma}
\affiliation{%
  \institution{Great Bay University}
  \city{Dongguan}
  \state{Guangdong}
  \country{China}
}

\author{Fei Ma}
\affiliation{%
  \institution{Guangdong Laboratory of Artificial Intelligence and Digital Economy (SZ)}
  \city{Shenzhen}
  \state{Guangdong}
  \country{China}
}

\author{Qi Tian}
\affiliation{%
  \institution{Guangdong Laboratory of Artificial Intelligence and Digital Economy (SZ)}
  \city{Shenzhen}
  \state{Guangdong}
  \country{China}
}

\renewcommand{\shortauthors}{Wang et al.}

\begin{document}

\begin{abstract}
LLM-based multimodal emotion recognition relies on static parametric memory and often hallucinates when interpreting nuanced affective states.
In this paper, given that single-round retrieval-augmented generation is highly susceptible to modal ambiguity and therefore struggles to capture complex affective dependencies across modalities, we introduce \textbf{AffectAgent}, an affect-oriented multi-agent retrieval-augmented generation framework that leverages collaborative decision-making among agents for fine-grained affective understanding.
Specifically, AffectAgent comprises three jointly optimized specialized agents, namely a query planner, an evidence filter, and an emotion generator, which collaboratively perform analytical reasoning to retrieve cross-modal samples, assess evidence, and generate predictions. These agents are optimized end-to-end using Multi-Agent Proximal Policy Optimization (MAPPO) with a shared affective reward to ensure consistent emotion understanding. 
Furthermore, we introduce Modality-Balancing Mixture of Experts (\textbf{MB-MoE}) and Retrieval-Augmented Adaptive Fusion (\textbf{RAAF}), where MB-MoE dynamically regulates the contributions of different modalities to mitigate representation mismatch caused by cross-modal heterogeneity, while RAAF enhances semantic completion under missing-modality conditions by incorporating retrieved audiovisual embeddings.
Extensive experiments on MER-UniBench demonstrate that AffectAgent achieves superior performance across complex scenarios. Our code will be released at: \url{https://github.com/Wz1h1NG1/AffectAgent}.
\end{abstract}






\maketitle
\newcommand{\introfigure}{%
\begin{figure}[t!]
\centering
\includegraphics[width=\linewidth]{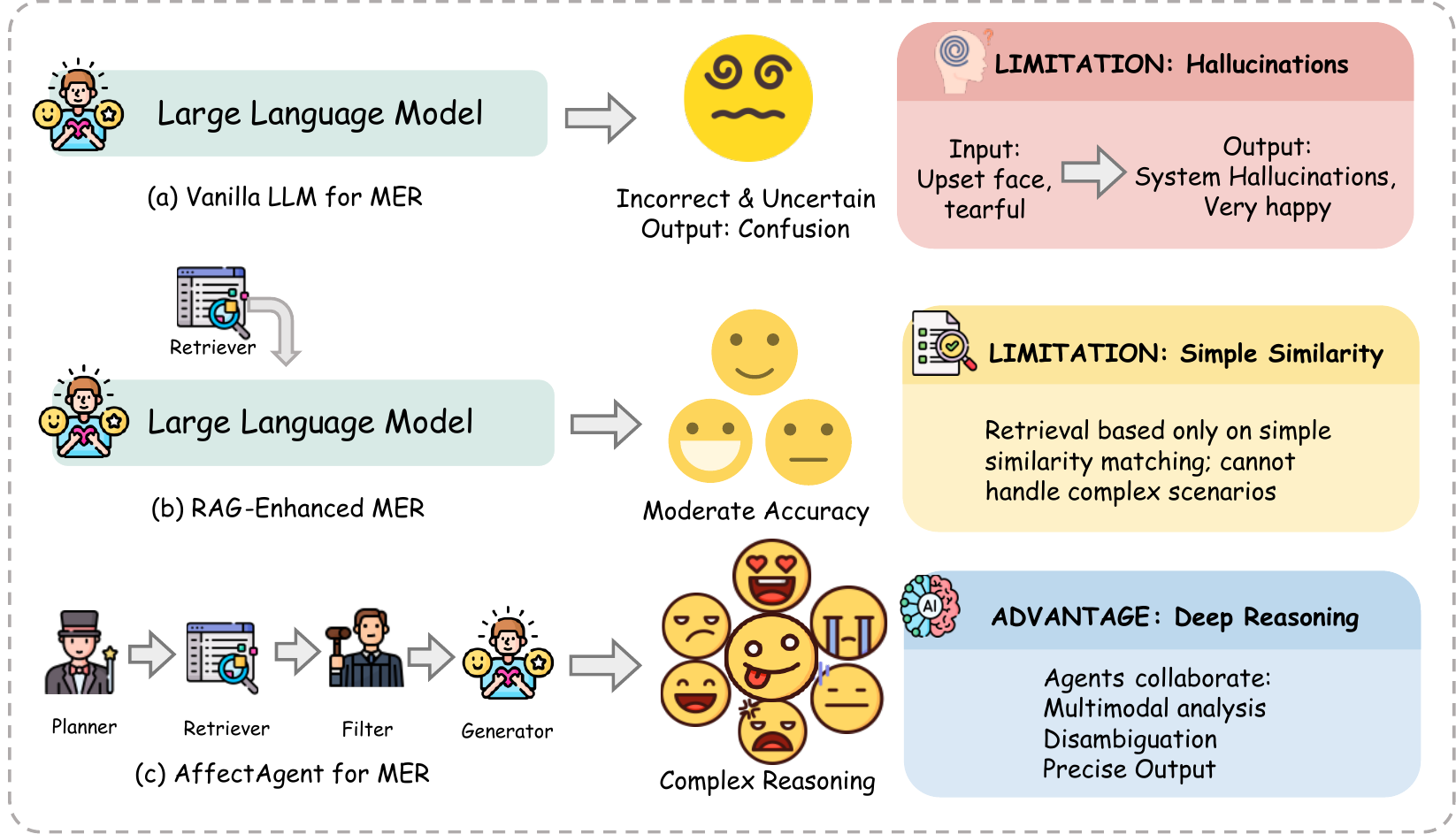}
\caption{
(a) \textbf{Vanilla LLMs}~\cite{lian2023affectgpt,cheng2024emotionllama} rely on static memory, struggling with nuanced emotion understanding.
(b) \textbf{Traditional RAG-enhanced models}~\cite{pipoli2025missrag,wen2025listen} follow a linear retrieve-then-generate pipeline, which is inadequate for complex emotional reasoning.
(c) \textbf{Our AffectAgent} adopts a collaborative multi-agent architecture that orchestrates query planning, retrieval, filtering, and generation for emotion recognition.
}
\label{fig:intro}
\vspace{-1em}
\end{figure}
}
\vspace{-0.7em}
\section{Introduction}
\textit{“Love is the one thing we’re capable of perceiving that transcends dimensions of time and space.”} Beyond its context in the film \textit{Interstellar}, this quote reflects a fundamental divide between algorithmic computation and human cognition: the perception of genuine emotion.
Human affect is inherently complex, and accurately perceiving and reasoning about human emotions is crucial for applications such as personalized education and psychological counseling \cite{soleymani2017survey, zadeh2018multimodal,xie2025physllm,zhao2026affectverse,zhao2026phase,huang2026complementarity,Zhao_2026_CVPR}.
Early unimodal methods \cite{li2022deep, wang2020region, schuller2018speech, el2011survey, devlin2019bert, lei2023instructerc, jia2022beyond} and multimodal fusion frameworks \cite{zadeh2017tensor, liu2018efficient, tsai2020multimodal, rahman2020integrating, han2021improving, zhang2023learning, cheng2023semi,zhu2025emosym,zhu2025uniemo,zhu2026H-GAR,zhu2026delta} struggle with this complexity, constrained by mechanical feature-level mappings lacking semantic understanding.

Recently, multimodal large language models have improved multimodal affective understanding by aligning different modalities in a shared linguistic space. They perform well on many vision-language tasks \cite{xue2025humanmotion,cheng2026splitgaussian,wang2026rsiccllm,wang2026dfm}, but their static parametric memory still limits their understanding of subtle emotions, as shown in Fig.~\ref{fig:intro}(a).
When confronted with severe cross-modal conflicts or dynamic conversational contexts, static pre-trained knowledge often defaults to ingrained language priors, leading to emotion hallucinations and compromising recognition accuracy \cite{deeppavlov2024acl}.
Multimodal retrieval-augmented generation addresses this problem by retrieving relevant cross-modal examples from external knowledge bases and grounding the model in dynamic evidence \cite{abootorabi2025ask, yuan2025mrag, pipoli2025missrag, wen2025listen,liu2025llm,liu2026aullmstructuralreasoninglarge,Wang2026MicroGesture,xin2024artificialintelligencecentraldogmacentric,xin2026hytrechybridtemporalawareattention,kong2026tokenreductionefficiencygenerative,Ma2026CLIPSA,Ma2026CMCLIP}, as shown in Fig.~\ref{fig:intro} (b).
However, conventional multimodal RAG relies on a single retrieval round. It is vulnerable to cross-modal ambiguity and struggles to model complex affective relations.

Collaborative multi-agent architectures \cite{wu2023autogen, liang2024multiagent}, such as MMOA \cite{chen2025mmoa}, employ multi-agent reinforcement learning to jointly optimize retrieval, filtering and generation, thereby addressing the limitations of isolated supervised fine-tuning \cite{gao2024modular, asai2024self}.
However, these joint optimization systems are primarily designed for objective, text-centric tasks. 

Human emotions are typically manifested through cross-modal cues. 
Existing multi-agent strategies primarily optimize for surface-level semantic relevance or explicit factual matching. 
If retrievers are trained to retrieve the semantically closest samples, they overlook the complex intrinsic cues that define real human affect. 
Without aligning specialized agents with the final emotion recognition goal, the retrieved samples fail to capture complex emotional cues.

In this paper, we propose AffectAgent, an affect-oriented multi-agent retrieval-augmented generation framework for fine-grained multimodal emotion recognition. It uses three agents that work together. The query planner generates supportive, confusing, and countering queries to retrieve different types of evidence. The evidence filter removes unreliable results, and the emotion generator combines the remaining evidence with the original input to predict the emotion.

\introfigure

We also introduce the Modality-Balancing Mixture of Experts, or MB-MoE, to handle modality imbalance. Retrieval-Augmented Adaptive Fusion, or RAAF, uses retrieved audiovisual embeddings to recover missing information. We train the agents and fusion modules together using Multi-Agent Proximal Policy Optimization, or MAPPO \cite{schulman2017proximal, yu2022surprising}. A shared affective reward ensures that the retrieved evidence supports the final prediction.
Our main contributions are as follows:
\begin{itemize}[leftmargin=2em, topsep=4pt, itemsep=4pt, parsep=0pt]
    \item 
	We propose \textbf{AffectAgent}, the first affect-oriented multi-agent retrieval-augmented generation framework for multimodal emotion recognition, which collaboratively reasons across modalities to capture nuanced affective states.	
    \item We introduce \textbf{MB-MoE} and \textbf{RAAF}: MB-MoE mitigates cross-modal mismatch by balancing modalities, while RAAF enhances semantic completion with retrieved audiovisual embeddings under missing modalities.
    \item Experiments on MER-UniBench \cite{lian2023affectgpt} show  AffectAgent generalizes across complex emotional scenarios.
\end{itemize}

\begin{figure*}[t]
    \centering
    \includegraphics[width=1\textwidth]{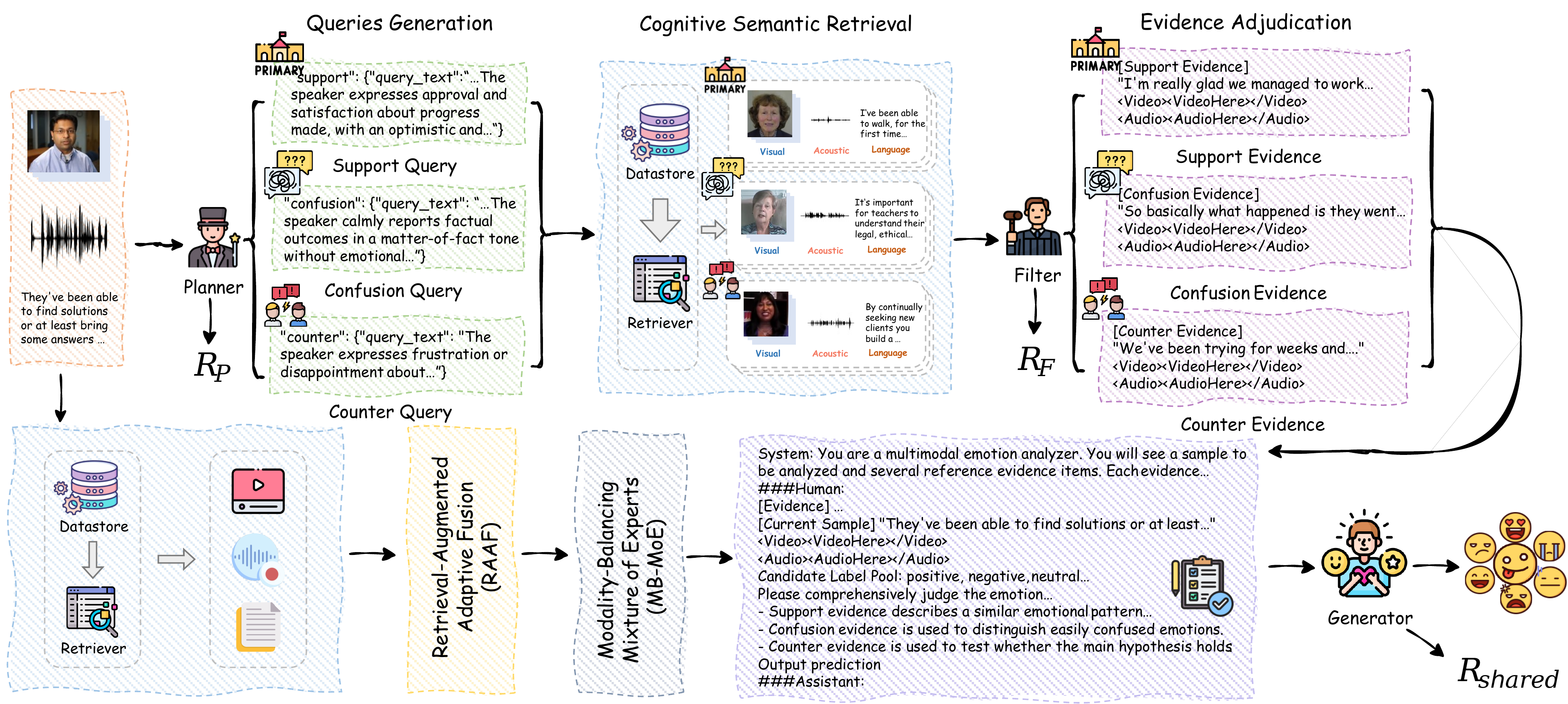}
    \caption{Overview of AffectAgent. AffectAgent performs multimodal emotion recognition through collaborative multi-agent reasoning, including affect-oriented query planning, evidence adjudication, and emotion generation. Meanwhile, RAAF and MB-MoE improve cross-modal fusion by recovering missing information and balancing modality contributions.}
    \Description{Block diagram of AffectAgent. Multimodal inputs are processed by query-planning, evidence-adjudication, and emotion-generation agents, while RAAF recovers missing information and MB-MoE balances the contributions of available modalities.}
    \label{fig:framework}
    \vspace{-1em}
\end{figure*}
\vspace{-1em}
\section{Related Works}
{\flushleft \textbf{Multimodal Emotion Recognition.}} MER \cite{ma2025generative} has evolved from feature alignment \cite{zadeh2017tensor, tsai2020multimodal} to MLLMs \cite{zhu2024minigpt, zhang2023videollama,chen2026reflectr1} and affective instruction tuning \cite{lian2023affectgpt, yang2024emollm,xing2026emollama,zhao2026affectverse,cheng2026omniopsd}. Yet static parametric knowledge and language priors remain unreliable for subtle or conflicting signals, causing emotion hallucinations \cite{deeppavlov2024acl}. AffectAgent instead grounds recognition in retrieved cross-modal evidence.

\vspace{-0.5em}
{\flushleft \textbf{Multimodal Retrieval-Augmented Generation.}} M-RAG grounds generation with retrieved cross-modal evidence \cite{chen2022murag, yasunaga2023racm3, hu2023reveal}, including affective references \cite{abootorabi2025ask, pipoli2025missrag}. Beyond linear retrieve-and-read pipelines \cite{yuan2025mrag, jiang2023flare}, agentic methods add iterative planning and collaboration \cite{yao2023react, wu2023autogen, liang2024multiagent}, while Self-RAG, Modular RAG, and MMOA introduce reflection, routing, or reinforcement learning \cite{asai2024self, gao2024modular, chen2025mmoa}. These methods prioritize relevance and factuality rather than subtle affective cues; AffectAgent instead couples affect-oriented query planning, evidence adjudication, and generation with a shared affective reward.

\vspace{-1em}
\section{Proposed Method}
\subsection{Overview of the framework}\label{sec:arch}

In this work, we propose AffectAgent, a multimodal retrieval-augmented generation framework for emotion understanding built on a collaborative multi-agent paradigm, where multiple components work in tandem. At each interactive step, the trainable components make joint decisions based on the current multimodal state to produce the final emotion recognition result. To coordinate these components for emotion understanding, we optimize the entire framework using Multi-Agent Proximal Policy Optimization (MAPPO). Rather than relying on disjoint local supervisory signals, MAPPO maximizes a shared affective objective across the whole system. This shared global reward enforces strict collaboration among components, ensuring that intermediate retrieval, filtering, and fusion operations genuinely support deep emotion understanding and preventing the model from degenerating into superficial lexical matching. As illustrated in Fig.~\ref{fig:framework}, AffectAgent consists of five core modules: a Query Planner, an Evidence Filter, an Emotion Generator, Modality-Balancing Mixture of Experts (MB-MoE), and Retrieval-Augmented Adaptive Fusion (RAAF):

\begin{itemize}[leftmargin=1em, topsep=4pt, itemsep=4pt, parsep=0pt]
    \item \textbf{Query Planner} analyzes the initial multimodal input $X_m$, where $m \in \{t, v, a\}$ denotes the text, visual, and audio modalities. Direct retrieval from $X_m$ only returns superficially similar evidence, which is insufficient for the Generator in complex emotional scenarios. The Planner therefore formulates a set of targeted cognitive hypotheses $Q = \{q_{sup}, q_{conf}, q_{count}\}$.
    
    \item \textbf{Retriever} fetches relevant multimodal evidence from external corpora and produces two outputs: cognitive evidence $E_{cog} = {E_{cog}^t, E_{cog}^v, E_{cog}^a}$ and perceptual evidence $E_{perc} = {E_{perc}^t, E_{perc}^v, E_{perc}^a}$, which are retrieved from the three formulated queries and from samples similar to the original input, respectively.
    
    \item \textbf{Evidence Filter} cross-verifies the cognitive evidence $E_{cog}$ against the original multimodal input to obtain a refined subset $E_{cog} = {E_{cog}^{t}, E_{cog}^{v}, E_{cog}^{a}}$ that is highly aligned with the input and thereby provides reliable support for final emotion recognition.
    
    \item \textbf{Modality-Balancing Mixture of Experts (MB-MoE)} balances cross-modal contributions to mitigate representation mismatch caused by cross-modal heterogeneity.
    
    \item \textbf{Retrieval-Augmented Adaptive Fusion (RAAF)} effectively combines retrieved perceptual embeddings with raw signals to enhance semantic completion, especially under missing-modality conditions.
    
    \item \textbf{Emotion Generator} leverages $E_{cog}$ together with the fused representation from RAAF and MB-MoE to produce the predicted emotion label $y$ and the explanatory rationale $r$ for the initial multimodal input.
\end{itemize}

We treat the Query Planner, Evidence Filter, and Emotion Generator as reinforcement learning agents. In this work, all three are implemented as multimodal large language models whose parameters are updated through reward signals. For computational efficiency, they share a single MLLM. In contrast, the Retriever involves a non-differentiable external search process and is therefore kept fixed as part of the environment. Unlike the Retriever, the MB-MoE and RAAF modules remain in the computation graph and are continuously updated through gradient backpropagation.

To align all agents and modules with the overall goal of genuine emotion understanding, we optimize them collaboratively. We define a shared reward function \(R_{shared}\) based on evaluation metrics of the Emotion Generator's output, such as F1 score, and apply it to all agents, following common practice in multi-agent reinforcement learning. To maintain the stability of each agent's specialized capability, we further design specific reward functions for the Planner and Filter. Detailed descriptions of these components are provided in the following subsections.

\vspace{-0.5em}
\subsection{Multi-Agent RL Formulation}\label{sec:reward_design}
We formulate AffectAgent as a partially observable multi-agent decision process. Since the agents sequentially operate on different modalities and intermediate artifacts, each agent observes only its local state rather than the full global state. Accordingly, the system is optimized end-to-end with multi-agent proximal policy optimization.

We define the shared reward by the F1 score between the Generator's predicted emotion label and the ground-truth label. Let $\hat{y}_{full}$ denote the prediction of the full system, $\hat{y}_{label}$ the prediction when the Query Planner output is replaced with a simple emotion label while the rest of the pipeline remains unchanged, and $\hat{y}_{rank}$ the prediction when the Evidence Filter is bypassed and the TopK retrieved items from each candidate group are directly passed to the Generator. The unified task scores are:
\vspace{-1.5em}

\begin{equation}
\text{Score}_{\star}=\text{F1}(\hat{y}_{\star}, y^*), \quad \star \in \{full, label, rank\}.
\vspace{-0.5em}
\end{equation}
All agents then jointly optimize the shared reward:
\vspace{-0.5em}
\begin{equation}
R_{shared} = \text{Score}_{full}.
\vspace{-0.25em}
\end{equation}
This means that all modules share the same optimization target: the final emotion prediction quality of the complete system. To provide clearer and timelier signals for agents other than the Generator, we introduce local incremental rewards in addition to the shared reward. Each local reward is defined by the gain in final F1 score over a simplified baseline, so that it reflects the contribution of the agent's specific function and is granted only when the agent truly improves the final performance.

{\flushleft \textbf{Environment.}} The environment includes the multimodal sample, the external evidence repository, the dual-pathway retriever, the modality-balancing mixture of experts (MB-MoE), the retrieval-augmented adaptive fusion module (RAAF), and the intermediate artifacts produced by the agents.

{\flushleft \textbf{State Transition.}} The global state evolves over three decision stages. It is initialized with the raw multimodal inputs and the candidate label set, then updated with the queries generated by the query planner and the candidate evidence returned by the retriever, and finally augmented with the refined evidence selected by the evidence filter and the fused multimodal representation produced by MB-MoE and RAAF.

{\flushleft \textbf{Observation Space.}} The agents observe different information across the three stages. In the first stage, the query planner observes the raw multimodal features $x_t$, $x_a$, and $x_v$, together with the candidate emotion label set $\mathcal{C}$. In the second stage, the evidence filter observes these raw features and the retrieved cognitive evidence $E_{cog}$. In the third stage, the emotion generator observes $x_t$, the refined cognitive evidence set $E_{cog}^*$, the fused multimodal representation from MB-MoE and RAAF, and the candidate label set $\mathcal{C}$.

{\flushleft \textbf{Action Space.}} All three agents act in the shared multimodal large language model vocabulary $\mathcal{V}$, but with different output formats. The query planner generates a token sequence $a_P = [a_P^1, \dots, a_P^{T_P}]$ ($a_P^j \in \mathcal{V}$) representing $q_{sup}$, $q_{conf}$, and $q_{count}$. For the $K$ items in $E_{cog}$, the evidence filter outputs a decision sequence $a_F = [a_F^1, \dots, a_F^K]$ ($a_F^i \in \{\text{Yes}, \text{No}\}$) to obtain $E_{cog}^*$. The emotion generator produces a final token sequence $a_G = [a_G^1, \dots, a_G^{T_G}]$ ($a_G^j \in \mathcal{V}$) as the rationale $r$ and the predicted label $\hat{y}_{full}$.

{\flushleft \textbf{Reward Functions.}} The emotion generator is optimized only by the shared reward, i.e., $R_G = R_{shared}$. To provide clearer signals for upstream agents, we further define local incremental rewards. For the query planner, we compare $\text{Score}_{full}$ with $\text{Score}_{label}$, yielding:
\begin{equation}
R_P = R_{shared} + \lambda_P \cdot (\text{Score}_{full} - \text{Score}_{label})
\end{equation}
For the evidence filter, we compare $\text{Score}_{full}$ with $\text{Score}_{rank}$, yielding
\begin{equation}
R_F = R_{shared} + \lambda_F \cdot (\text{Score}_{full} - \text{Score}_{rank})
\end{equation}

\vspace{-1em}
\subsection{Retrieval-Augmented Adaptive Fusion}
To address the issues of modality imbalance and modality missing, AffectAgent further exploits the perceptual evidence \(E_{perc}\) from the retrieval branch. To avoid interfering with the cognitive evidence selection, we only enhance the audiovisual features. Specifically, given the sample features \(x_v\) and \(x_a\), we denote the visual and audio features of the retrieved support sample as \(E_{perc}^v\) and \(E_{perc}^a\), respectively. Retrieval-Augmented Adaptive Fusion (RAAF) then enhances them through gated cross-attention. For modality \(m\in\{v,a\}\), the fusion process is defined as
\begin{equation}
\begin{aligned}
h_m &= \mathrm{Attn}(x_m,E_{perc}^m,E_{perc}^m),
\end{aligned}
\end{equation}
\begin{equation}
\begin{aligned}
\hat{x}_m &= x_m+\sigma\!\left(W_m[x_m;h_m]\right)\odot h_m.
\end{aligned}
\end{equation}
where \([\cdot;\cdot]\) denotes concatenation and \(\odot\) denotes element-wise multiplication. In this way, RAAF injects only useful complementary cues from the retrieved support sample while preserving the dominant content of the current input.

\vspace{-1em}
\subsection{Modality-Balancing MoE}

Although RAAF enhances the perceptual representation, the relative reliability of video and audio may still vary across samples due to modality imbalance. To address this issue, we introduce a Modality-Balancing Mixture of Experts (MB-MoE), which adaptively refines the two modalities based on the overall audiovisual information.

Specifically, MB-MoE first pools the RAAF-enhanced features \(\hat{x}_v\) and \(\hat{x}_a\) into a global state \(g\). A router then takes \(g\) to compute expert scores, selects the top-\(K\) experts \(\mathcal{K}\), and applies softmax function to obtain the normalized weights \(\alpha\). Finally, the selected experts are applied to both modalities:
\vspace{-0.5em}
\begin{equation}
\tilde{x}_m = \sum_{j\in\mathcal{K}} \alpha_j f_j(\hat{x}_m), \qquad m\in\{v,a\},
\vspace{-0.5em}
\end{equation}

where \(f_j(\cdot)\) denotes the \(j\)-th expert transformation. By sharing the same routing weights \(\alpha\) derived from the global state \(g\), expert selection is coordinated across modalities. Rather than simply assigning scalar weights to video and audio, this shared mechanism applies the most suitable feature transformations to pull both modalities into a coordinated and balanced representation space. The final fused multimodal representation is denoted by \(z_{fuse}=\{\tilde{x}_v,\tilde{x}_a\}\). The Emotion Generator then conditions on the raw text input \(x_t\), the refined cognitive evidence \(E'_{cog}\), and \(z_{fuse}\) to produce the final emotion label and rationale.

\vspace{-1em}
\subsection{Multi-Agent Optimization}

Before joint optimization, we use role-specific supervised fine-tuning \cite{huang2026preprompt,huang2024etag,wang2024revisiting,xing2025knowledge,wu2025navigating,he2023mitigating,he2024progressive} to initialize the Query Planner, Evidence Filter, and Emotion Generator. The three agents learn query generation, evidence selection, and emotion prediction, respectively. Further details of this warm-start procedure are provided in the appendix. We then model AffectAgent as a fully cooperative system and apply a joint training strategy based on Multi-Agent PPO, or MAPPO \cite{yu2022surprising}, to improve agent collaboration. Following Sections~\ref{sec:arch} and~\ref{sec:reward_design}, the agents share a single MLLM to reduce computation. The Retriever remains fixed because it is non-differentiable, while MB-MoE and RAAF are jointly updated with the shared MLLM.

In the multi-agent optimization process, three models are considered: the Actor, the Critic, and the frozen SFT model, whose parameters are denoted by \(\theta\), \(\phi\), and \(\theta_{\mathrm{SFT}}\), respectively. For each agent \(i\in\{Q,F,G\}\), the Actor is responsible for providing the collaborative decision policy (generating the complete output sequence \(A_i\)) based on the observation \(o_i\). The Critic model estimates the state-value function to stabilize the optimization process, a classic setup in the Actor-Critic architecture within RL algorithms. The SFT model serves as a reference baseline for the Actor model for KL regularization \cite{ouyang2022training,wang2024improving,chen2026robust}. The overall objective to update the parameters of both the Actor and Critic models, \(\mathcal{L}(\theta,\phi)\), consists of two terms, where \(\alpha\) is the coefficient balancing policy optimization and value fitting:
\vspace{-1em}

\begin{equation}
\mathcal{L}(\theta,\phi)=\mathcal{L}_{\mathrm{Actor}}(\theta)+\alpha\,\mathcal{L}_{\mathrm{Critic}}(\phi)
\end{equation}


The Actor loss is similar to that used in the standard single-agent PPO algorithm \cite{schulman2017proximal}, with the main difference that, in our setting, multiple agents are optimized jointly. Specifically, \(i \in \{Q,F,G\}\) denotes the three agents considered in our framework. For each agent, \(r_t^i\) denotes the importance sampling ratio, which measures the discrepancy between the updated policy and the old rollout policy parameterized by \(\theta_{\mathrm{old}}\). As in PPO, this ratio is introduced to constrain policy updates and improve training stability.

The term \(\hat{A}_t^i\) denotes the advantage estimate for agent \(i\), which is computed using Generalized Advantage Estimation (GAE) \cite{schulman2016high}. This estimation provides a practical balance between bias and variance in policy gradient optimization. In addition, \(\delta_t^i\) represents the temporal-difference (TD) error at time step \(t\), which is used to recursively compute the advantage. Here, \(\gamma\) and \(\lambda\) denote the discount factor and the GAE coefficient, respectively:


\begin{equation}
\left\{
\begin{aligned}
\mathcal{L}_{\mathrm{Actor}}(\theta) &= - \sum_i \sum_t \min\!\left( r_t^i\hat{A}_t^i,\; \operatorname{clip}(r_t^i,1-\epsilon,1+\epsilon)\hat{A}_t^i \right) \\
r_t^i &= \frac{\pi_{\theta}(a_t^i\mid s_t^i)}{\pi_{\theta_{\mathrm{old}}}(a_t^i\mid s_t^i)} \\
\hat{A}_t^i &= \delta_t^i+\gamma\lambda \hat{A}_{t+1}^i \\
\delta_t^i &= R(s_t^i,a_t^i)+\gamma V_{\phi}(s_{t+1}^i)-V_{\phi}(s_t^i)
\end{aligned}
\right.
\end{equation}

Similar to InstructGPT \cite{ouyang2022training}, the final reward function \(R(s_t^i,a_t^i)\) incorporates a sequence-level KL regularization penalty. The distinction is that our approach does not require an additionally trained reward model; instead, we directly use the predefined task rewards \(R_i\) introduced in Section~\ref{sec:reward_design}. Since reliable supervision becomes available only after the complete collaborative process is finished, this terminal reward is assigned only at the end of the trajectory (i.e., \(t=T_i\)). Here, \(\beta\) controls the strength of the KL regularization to keep the updated policy close to the SFT initialization:
\begin{equation}
R(s_t^i,a_t^i)=
\begin{cases}
0, & \text{if } t<T_i,\\
R_i-\beta \log \dfrac{\pi_{\theta_{\mathrm{old}}}(A_i\mid o_i)}{\pi_{\theta_{\mathrm{SFT}}}(A_i\mid o_i)}, & \text{if } t=T_i.
\end{cases}
\end{equation}


The Critic loss also adopts a clipping strategy similar to that used in the Actor objective, with the goal of stabilizing value function optimization and preventing excessively large updates between successive training iterations. Specifically, we define the value target as \(V_{\mathrm{target}}^{i,t}=\hat{A}_t^i+V_{\phi}(s_t^i)\), where \(\hat{A}_t^i\) is the estimated advantage and \(V_{\phi}(s_t^i)\) is the current value prediction. To further improve training stability, we introduce a clipped value estimate
\begin{equation}
\tilde{V}_{\phi}(s_t^i)
=
\operatorname{clip}\!\left(
V_{\phi}(s_t^i),\;
V_{\phi_{\mathrm{old}}}(s_t^i)-\epsilon,\;
V_{\phi_{\mathrm{old}}}(s_t^i)+\epsilon
\right),
\end{equation}
where the clipping range is centered at the prediction of the old value network parameterized by \(\phi_{\mathrm{old}}\). This design constrains the change in value prediction within a local trust region and makes the regression objective more robust during training:

\begin{equation}
\mathcal{L}_{\mathrm{Critic}}(\phi)
=
\sum_i \sum_t
\max \Biggl\{
\begin{array}{l}
\left( V_{\phi}(s_t^i) - V_{\mathrm{target}}^{i,t} \right)^2, \\[4pt]
\left( \tilde{V}_{\phi}(s_t^i) - V_{\mathrm{target}}^{i,t} \right)^2
\end{array}
\Biggr\}.
\end{equation}

Each training iteration begins with a complete rollout. The data passes through the Query Planner, Retriever, Evidence Filter, MB-MoE, RAAF, and Emotion Generator, and the resulting trajectories are stored in the replay buffer. We then compute rewards, advantages, and value targets for the three agents. Several PPO epochs update the Actor and Critic parameters in parallel. After training, we discard the Critic and the SFT reference model. Inference and evaluation use only the optimized shared Actor and the updated MB-MoE and RAAF modules.

\definecolor{mygray}{gray}{0.88} 
\definecolor{upcolor}{gray}{0.4} 

\begin{table*}[t]
\centering
\caption{Results on \textbf{MER-UniBench}. For each dataset, we report its primary evaluation metric. The gray-shaded columns denote the primary ranking metrics, the best results are highlighted in bold, and the second-best results are underlined. Our \textit{AffectAgent} consistently outperforms different baseline models.}
\vspace{-1em}
\newcommand{\up}[1]{\;{\scriptsize\color{upcolor}{\textbf{$\uparrow$}#1}}} 
\resizebox{\textwidth}{!}{
\begin{tabular}{l ccc | cccc | cccc | c | c}
\toprule
\multirow{2}{*}{\textbf{Model}} & \multicolumn{3}{c|}{\textbf{Modality}} & \multicolumn{4}{c|}{\textbf{Basic}} & \multicolumn{4}{c|}{\textbf{Sentiment}} & \textbf{Fine-grained} & \multirow{2}{*}{\textbf{Mean}} \\
 & A & V & T & MER2023 & MER2024 & MELD & IEMOCAP & MOSI & MOSEI & SIMS & SIMS v2 & OV-MERD+ & \\
\midrule
Video-LLaMA~\cite{zhang2023videollama}     & $\checkmark$ & $\checkmark$ & $\checkmark$ & 14.52 & 13.56 & 17.65 & 15.68 & 54.55 & 46.52 & 58.56 & 56.32 & 13.98 & 32.37 \\
\rowcolor{mygray} \quad \textit{+ AffectAgent} & & & & 22.75\up{8.23} & 22.71\up{9.15} & 24.07\up{6.42} & 23.49\up{7.81} & 58.90\up{4.35} & 52.14\up{5.62} & 61.70\up{3.14} & 60.40\up{4.08} & 22.65\up{8.67} & 38.75\up{6.38} \\
\midrule
VideoChat~\cite{li2024videochat}           & $\checkmark$ & $\checkmark$ & $\checkmark$ & 16.89 & 15.65 & 19.55 & 16.98 & 56.02 & 48.65 & 59.68 & 57.65 & 15.02 & 34.01 \\
\rowcolor{mygray} \quad \textit{+ AffectAgent} & & & & 24.30\up{7.41} & 23.87\up{8.22} & 25.38\up{5.83} & 23.93\up{6.95} & 61.14\up{5.12} & 54.96\up{6.31} & 63.93\up{4.25} & 61.53\up{3.88} & 22.36\up{7.34} & 40.15\up{6.14} \\
\midrule
ChatBridge~\cite{zhao2023chatbridge}       & $\checkmark$ & $\checkmark$ & $\checkmark$ & 18.06 & 16.89 & 20.35 & 25.02 & 56.66 & 49.68 & 58.62 & 56.55 & 15.35 & 35.24 \\
\rowcolor{mygray} \quad \textit{+ AffectAgent} & & & & 27.18\up{9.12} & 24.54\up{7.65} & 28.39\up{8.04} & 30.23\up{5.21} & 61.53\up{4.87} & 55.60\up{5.92} & 64.73\up{6.11} & 62.29\up{5.74} & 23.80\up{8.45} & 42.03\up{6.79} \\
\midrule
Video-LLaMA 2~\cite{cheng2024videollama}   & $\checkmark$ & $\checkmark$ & $\checkmark$ & 20.24 & 18.32 & 22.08 & 23.32 & 54.38 & 50.12 & 60.23 & 58.69 & 16.58 & 35.99 \\
\rowcolor{mygray} \quad \textit{+ AffectAgent} & & & & 28.80\up{8.56} & 27.53\up{9.21} & 29.41\up{7.33} & 29.80\up{6.48} & 60.13\up{5.75} & 56.26\up{6.14} & 64.75\up{4.52} & 63.60\up{4.91} & 24.44\up{7.86} & 42.74\up{6.75} \\
\midrule
PandaGPT~\cite{su2023pandagpt}             & $\checkmark$ & $\checkmark$ & $\checkmark$ & 40.21 & 51.89 & 37.88 & 44.04 & 61.92 & 67.61 & 68.38 & 67.23 & 37.12 & 52.92 \\
\rowcolor{mygray} \quad \textit{+ AffectAgent} & & & & 46.33\up{6.12} & 56.24\up{4.35} & 45.29\up{7.41} & 49.92\up{5.88} & 65.16\up{3.24} & 70.56\up{2.95} & 70.52\up{2.14} & 70.31\up{3.08} & 43.67\up{6.55} & 58.66\up{5.74} \\
\midrule
Emotion-LLaMA~\cite{cheng2024emotionllama} & $\checkmark$ & $\checkmark$ & $\checkmark$ & 59.38 & 73.62 & 46.76 & 55.47 & 66.13 & 67.66 & 78.32 & 77.23 & 52.97 & 64.17 \\
\rowcolor{mygray} \quad \textit{+ AffectAgent} & & & & 63.59\up{4.21} & 76.46\up{2.84} & 52.38\up{5.62} & 59.80\up{4.33} & 69.54\up{3.41} & 71.54\up{3.88} & 80.27\up{1.95} & 79.35\up{2.12} & 57.72\up{4.75} & 67.85\up{3.68} \\
\midrule
AffectGPT~\cite{lian2023affectgpt}         & $\checkmark$ & $\checkmark$ & $\checkmark$ & \underline{78.54} & \underline{78.80} & \underline{55.65} & \underline{60.54} & \underline{81.30} & \underline{80.90} & \underline{88.49} & \underline{86.18} & \underline{62.52} & \underline{74.77} \\
\rowcolor{mygray} \textbf{\quad \textit{+ AffectAgent}} & & & & \textbf{80.69}\up{2.15} & \textbf{80.66}\up{1.86} & \textbf{59.07}\up{3.42} & \textbf{63.39}\up{2.85} & \textbf{82.73}\up{1.43} & \textbf{82.67}\up{1.77} & \textbf{89.43}\up{0.94} & \textbf{87.33}\up{1.15} & \textbf{65.13}\up{2.61} & \textbf{76.78}\up{2.01} \\
\bottomrule
\end{tabular}
}
\label{tab:full_modality_marag}
\vspace{-1em}
\end{table*}
\begin{table*}[ht]
\centering
\caption{Robustness evaluation under missing-modality settings. \textbf{A}, \textbf{V}, and \textbf{T} denote audio, video, and text, respectively. Results show that \textit{AffectAgent} consistently improves performance across different backbone models under various missing-modality conditions.}
\vspace{-1em}
\newcommand{\up}[1]{\;{\scriptsize\color{upcolor}{\textbf{$\uparrow$}#1}}} 
\resizebox{\textwidth}{!}{
\begin{tabular}{l ccc | cccc | cccc | c | c}
\toprule
\multirow{2}{*}{\textbf{Model}} & \multicolumn{3}{c|}{\textbf{Modality}} & \multicolumn{4}{c|}{\textbf{Basic}} & \multicolumn{4}{c|}{\textbf{Sentiment}} & \textbf{Fine-grained} & \multirow{2}{*}{\textbf{Mean}} \\
 & A & V & T & MER2023 & MER2024 & MELD & IEMOCAP & MOSI & MOSEI & SIMS & SIMS v2 & OV-MERD+ & \\
\midrule
PandaGPT~\cite{su2023pandagpt} & $\checkmark$ & $\checkmark$ & $\times$ & 28.15 & 32.40 & 26.50 & 31.20 & 55.30 & 52.10 & 54.20 & 50.50 & 25.40 & 39.53 \\
\rowcolor{mygray} \quad \textit{+ AffectAgent} & & & & 32.27\up{4.12} & 35.95\up{3.55} & 31.71\up{5.21} & 36.03\up{4.83} & 57.45\up{2.15} & 54.94\up{2.84} & 56.12\up{1.92} & 52.91\up{2.41} & 29.76\up{4.36} & 43.01\up{3.48} \\
Emotion-LLaMA~\cite{cheng2024emotionllama} & $\checkmark$ & $\checkmark$ & $\times$ & 45.20 & 56.80 & 35.40 & 42.10 & 56.50 & 58.20 & 65.30 & 64.10 & 40.50 & 51.57 \\
\rowcolor{mygray} \quad \textit{+ AffectAgent} & & & & 49.01\up{3.81} & 58.94\up{2.14} & 39.52\up{4.12} & 45.66\up{3.56} & 58.73\up{2.23} & 60.65\up{2.45} & 67.16\up{1.86} & 66.02\up{1.92} & 44.35\up{3.85} & 54.44\up{2.87} \\
AffectGPT~\cite{lian2023affectgpt} & $\checkmark$ & $\checkmark$ & $\times$ & \underline{65.20} & \underline{66.80} & \underline{48.50} & \underline{52.40} & \underline{72.10} & \underline{70.30} & \underline{78.40} & \underline{76.50} & \underline{52.60} & \underline{64.76} \\
\rowcolor{mygray} \textbf{\quad \textit{+ AffectAgent}} & & & & \textbf{68.05}\up{2.85} & \textbf{68.92}\up{2.12} & \textbf{51.64}\up{3.14} & \textbf{55.15}\up{2.75} & \textbf{74.02}\up{1.92} & \textbf{72.15}\up{1.85} & \textbf{79.81}\up{1.41} & \textbf{78.02}\up{1.52} & \textbf{55.53}\up{2.93} & \textbf{67.03}\up{2.27} \\
\midrule
PandaGPT~\cite{su2023pandagpt} & $\checkmark$ & $\times$ & $\checkmark$ & 33.57 & 39.04 & 31.91 & 36.55 & 66.06 & 61.33 & 62.93 & 58.88 & 31.33 & 46.84 \\
\rowcolor{mygray} \quad \textit{+ AffectAgent} & & & & 37.72\up{4.15} & 42.46\up{3.42} & 35.76\up{3.85} & 40.67\up{4.12} & 66.71\up{0.65} & 63.17\up{1.84} & 65.08\up{2.15} & 60.93\up{2.05} & 35.01\up{3.68} & 49.72\up{2.88} \\
Emotion-LLaMA~\cite{cheng2024emotionllama} & $\checkmark$ & $\times$ & $\checkmark$ & 52.15 & 64.30 & 41.25 & 49.60 & 62.40 & 63.80 & 72.15 & 70.40 & 48.20 & 58.25 \\
\rowcolor{mygray} \quad \textit{+ AffectAgent} & & & & 55.56\up{3.41} & 67.15\up{2.85} & 44.89\up{3.64} & 52.82\up{3.22} & 64.55\up{2.15} & 65.88\up{2.08} & 74.07\up{1.92} & 72.54\up{2.14} & 51.55\up{3.35} & 61.00\up{2.75} \\
AffectGPT~\cite{lian2023affectgpt} & $\checkmark$ & $\times$ & $\checkmark$ & \underline{72.94} & \underline{73.41} & \underline{56.63} & \underline{55.68} & \underline{83.46} & \underline{80.74} & \underline{82.99} & \underline{83.75} & \underline{59.98} & \underline{72.18} \\
\rowcolor{mygray} \textbf{\quad \textit{+ AffectAgent}} & & & & \textbf{74.79}\up{1.85} & \textbf{74.93}\up{1.52} & \textbf{57.87}\up{1.24} & \textbf{57.83}\up{2.15} & \textbf{83.81}\up{0.35} & \textbf{81.86}\up{1.12} & \textbf{84.44}\up{1.45} & \textbf{84.93}\up{1.18} & \textbf{62.02}\up{2.04} & \textbf{73.61}\up{1.43} \\
\midrule
PandaGPT~\cite{su2023pandagpt} & $\times$ & $\checkmark$ & $\checkmark$ & 39.13 & 47.16 & 38.33 & 47.21 & 58.50 & 64.25 & 62.07 & 65.25 & 35.07 & 50.77 \\
\rowcolor{mygray} \quad \textit{+ AffectAgent} & & & & 42.98\up{3.85} & 50.28\up{3.12} & 41.98\up{3.65} & 49.75\up{2.54} & 61.35\up{2.85} & 66.86\up{2.61} & 64.41\up{2.34} & 67.47\up{2.22} & 39.22\up{4.15} & 53.81\up{3.04} \\
Emotion-LLaMA~\cite{cheng2024emotionllama} & $\times$ & $\checkmark$ & $\checkmark$ & 55.40 & 68.10 & 44.50 & 52.30 & 64.50 & 65.20 & 75.60 & 74.10 & 50.80 & 61.17 \\
\rowcolor{mygray} \quad \textit{+ AffectAgent} & & & & 58.35\up{2.95} & 70.44\up{2.34} & 47.65\up{3.15} & 55.12\up{2.82} & 66.85\up{2.35} & 67.71\up{2.51} & 77.44\up{1.84} & 76.05\up{1.95} & 53.92\up{3.12} & 63.73\up{2.56} \\
AffectGPT~\cite{lian2023affectgpt} & $\times$ & $\checkmark$ & $\checkmark$ & \underline{74.58} & \underline{75.29} & \underline{57.63} & \underline{62.19} & \underline{82.39} & \underline{81.57} & \underline{87.20} & \underline{86.29} & \underline{61.65} & \underline{74.31} \\
\rowcolor{mygray} \textbf{\quad \textit{+ AffectAgent}} & & & & \textbf{76.23}\up{1.65} & \textbf{76.71}\up{1.42} & \textbf{58.78}\up{1.15} & \textbf{63.27}\up{1.08} & \textbf{82.87}\up{0.48} & \textbf{82.42}\up{0.85} & \textbf{88.42}\up{1.22} & \textbf{87.23}\up{0.94} & \textbf{63.50}\up{1.85} & \textbf{75.49}\up{1.18} \\
\bottomrule
\end{tabular}
}
\label{tab:grouped_modality_ablation}
\end{table*}

\begin{table*}[h]
\centering
\caption{Performance comparison of different RAG methods on emotion recognition datasets. To ensure a fair comparison, all RAG methods are applied to the same MLLM using full-modality inputs (audio, video, and text). Our proposed \textit{AffectAgent} consistently outperforms existing RAG baselines across all datasets.}
\vspace{-1em}
\newcommand{\up}[1]{\;{\scriptsize\color{upcolor}{\textbf{$\uparrow$}#1}}} 
\resizebox{\textwidth}{!}{
\begin{tabular}{l | cccc | cccc | c | c}
\toprule
\multirow{2}{*}{\textbf{Method}} & \multicolumn{4}{c|}{\textbf{Basic}} & \multicolumn{4}{c|}{\textbf{Sentiment}} & \textbf{Fine-grained} & \multirow{2}{*}{\textbf{Mean}} \\
 & MER2023 & MER2024 & MELD & IEMOCAP & MOSI & MOSEI & SIMS & SIMS v2 & OV-MERD+ & \\
\midrule
No RAG                  & 78.54 & 78.80 & 55.65 & 60.54 & 81.30 & 80.90 & 88.49 & 86.18 & 62.52 & 74.77 \\
Rewrite-Retrieve-Read~\cite{ma2023query} & 78.39 & 78.76 & 55.58 & 60.42 & 81.36 & 80.83 & 88.52 & 86.13 & 62.47 & 74.72 \\
BGM~\cite{ke2024bridging}                    & 78.68 & 79.03 & 55.88 & 60.77 & 81.62 & 81.18 & 88.73 & 86.41 & 62.81 & 75.01 \\
RAG-DDR~\cite{li2025ragddr}                & \underline{79.12} & \underline{79.38} & \underline{56.33} & \underline{61.27} & \underline{81.98} & \underline{81.54} & \underline{89.13} & \underline{86.82} & \underline{63.21} & \underline{75.42} \\
\rowcolor{mygray} \textbf{AffectAgent (Ours)} & \textbf{80.69}\up{2.15} & \textbf{80.66}\up{1.86} & \textbf{59.07}\up{3.42} & \textbf{63.39}\up{2.85} & \textbf{82.73}\up{1.43} & \textbf{82.67}\up{1.77} & \textbf{89.43}\up{0.94} & \textbf{87.33}\up{1.15} & \textbf{65.13}\up{2.61} & \textbf{76.78}\up{2.01} \\
\bottomrule
\end{tabular}
}
\vspace{-1em}
\label{tab:rag_method_comparison}
\end{table*}

\vspace{-1em}
\section{Experiments}
Our experiments evaluate AffectAgent across different MLLMs and missing-modality settings and compare it with existing RAG methods. We also validate its core designs and present parameter analyses and case studies.
\vspace{-1em}
\subsection{Experimental Setup}
\noindent\textbf{Datasets and Evaluation.} We evaluate our method on the \textbf{MER-UniBench} benchmark~\cite{lian2023affectgpt}, which covers three representative multimodal emotion understanding settings: Basic Emotion Recognition, Sentiment Analysis, and Fine-grained Emotion Recognition. Following the benchmark protocol, we report the primary metric for each dataset: \textbf{HIT} for MER2023, MER2024, MELD, and IEMOCAP; \textbf{WAF} for MOSI, MOSEI, SIMS, and SIMS v2; and \textbf{$F_s$} for OV-MERD+. In addition, we build a multimodal retrieval library from the MER2025~\cite{lian2023affectgpt} training set and jointly train on this dataset, which serves as the external evidence source for retrieval-augmented emotion reasoning. All methods are evaluated under the same protocol for fair comparison.
\begin{sloppypar}
\noindent\textbf{Implementation Details.} We use E5~\cite{wang2024multilingual} for text retrieval and a FAISS-based nearest-neighbor retriever~\cite{johnson2019billion} for multimodal retrieval. The selector receives \(K\) evidence samples as input. We evaluate seven MLLMs, including emotion-oriented models, i.e., AffectGPT~\cite{lian2023affectgpt} and Emotion-LLaMA~\cite{cheng2024emotionllama}, and general-purpose models, i.e., Video-LLaMA~\cite{zhang2023videollama}, Video-LLaMA2~\cite{cheng2024videollama}, VideoChat~\cite{li2024videochat}, ChatBridge\cite{zhao2023chatbridge}, and PandaGPT\cite{su2023pandagpt}. We further compare with three RAG baselines: Rewrite-Retrieve-Read~\cite{ma2023query}, BGM~\cite{ke2024bridging}, and RAG-DDR~\cite{li2025ragddr}.
\end{sloppypar}
\vspace{-0.8em}
\subsection{Main Results across Different MLLMs}

Table~\ref{tab:full_modality_marag} reports the full-modality results across different MLLM backbones. AffectAgent consistently improves all compared models on MER-UniBench, covering Basic Emotion Recognition, Sentiment Analysis, and Fine-grained Emotion Recognition. The gains are observed in both overall mean scores and most individual datasets, demonstrating its broad compatibility with diverse multimodal LLMs and effectiveness for multimodal emotion reasoning.

The improvements are pronounced on general-purpose MLLMs, including Video-LLaMA, VideoChat, ChatBridge, Video-LLaMA2 and PandaGPT. Compared with stronger backbones, these models benefit more from AffectAgent, with mean scores increasing from 32.37 to 38.75, 34.01 to 40.15, 35.24 to 42.03, 35.99 to 42.74, and 52.92 to 58.66, respectively. Similar gains are observed across Basic Emotion Recognition, Sentiment Analysis, and Fine-grained Emotion Recognition, indicating that AffectAgent is especially effective for general-purpose MLLMs with lower performance.

Stable gains are also observed on stronger affect-oriented backbones, including Emotion-LLaMA and AffectGPT. Although the absolute improvements are smaller on these models, the gains remain consistent across datasets and task settings. This indicates that AffectAgent is not only useful for weaker backbones, but can also provide additional benefits for models that already have strong affective knowledge. Moreover, the improvements are observed across all three task groups in MER-UniBench, rather than being limited to a single benchmark. This suggests that the proposed framework generalizes well across affective tasks with different label granularity and reasoning difficulty, and works as a generally effective retrieval-augmented reasoning framework rather than a backbone-specific add-on.
\vspace{-1em}
\begin{figure*}[!t]
	\centering
	\includegraphics[width=\textwidth]{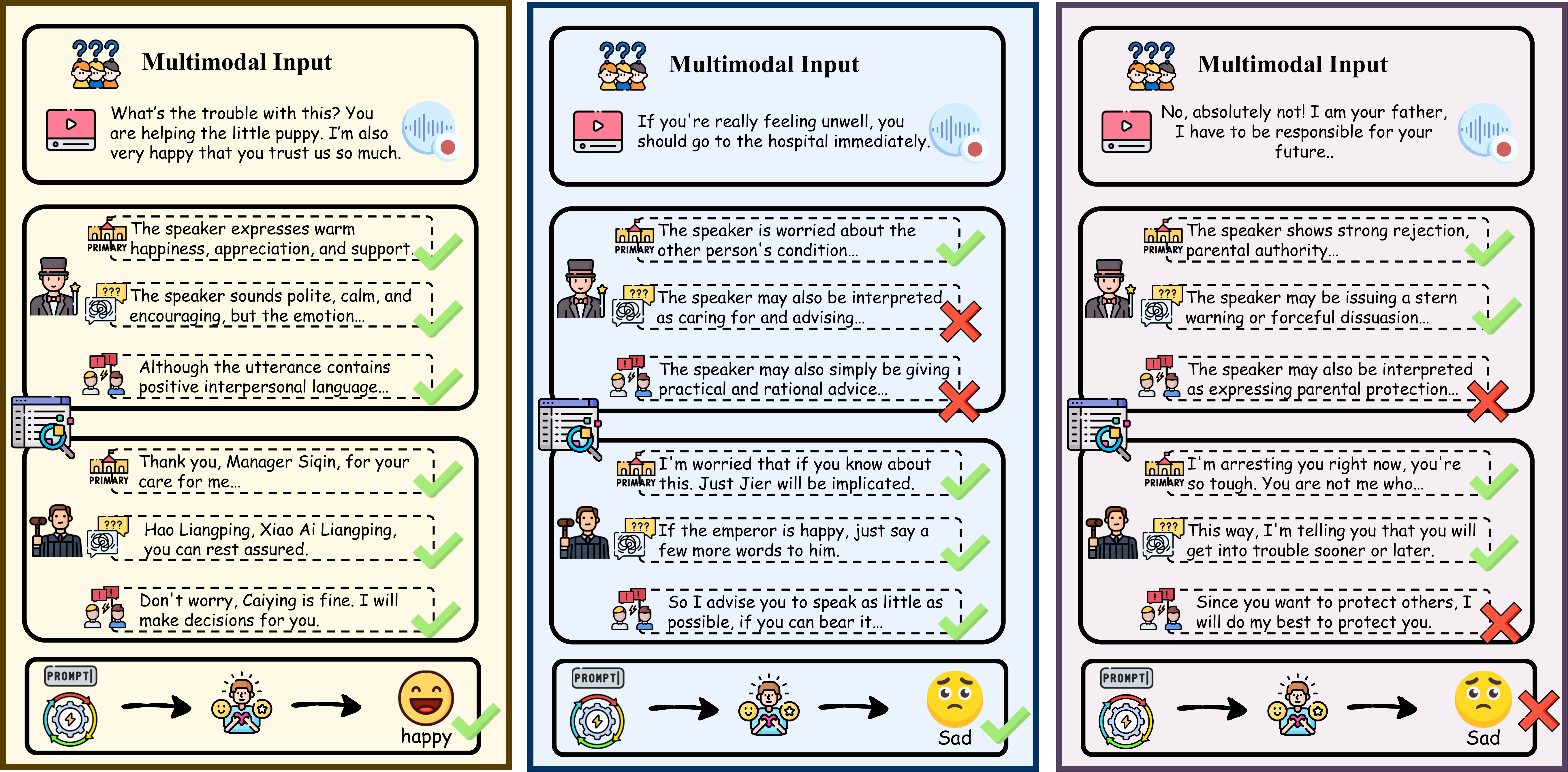}
	\caption{Illustration of three representative case studies, showing smooth success, query deviation with a correct final prediction, and query deviation with an incorrect final prediction.}
	\Description{Three qualitative examples trace AffectAgent from its generated query through retrieved evidence and reasoning to the final emotion prediction. The examples show a smooth success, a deviated query with a correct prediction, and a deviated query with an incorrect prediction.}
	\label{fig:case_study}
\end{figure*}
\subsection{Robustness to Missing Modalities}
Table~\ref{tab:grouped_modality_ablation} presents the experimental results under missing-modality settings. When a modality is missing, we directly supplement it with retrieved multimodal perceptual evidence. Overall, AffectAgent consistently improves the performance of all compared models across different modality combinations, including \textit{A+V}, \textit{A+T}, and \textit{V+T}. This demonstrates that the proposed framework remains effective when a single modality is unavailable, showing strong robustness to incomplete multimodal inputs.

We observe that the improvement margins vary across modality combinations. In general, larger gains occur when the missing modality leads to a more severe loss of emotional evidence, while stable improvements are still maintained under relatively stronger modality combinations. These results indicate that AffectAgent not only enhances performance in the full-modality setting, but also improves the robustness of MLLMs under missing-modality conditions.
\begin{table}[h]
\centering
\caption{Stepwise effectiveness of our components. We utilize the generic MLLM \textbf{Video-LLaMA2} as the base model. $\Delta$HIT indicates the absolute improvement over the zero-shot baseline.}
\label{tab:main_results}
\vspace{-1em}
\renewcommand{\arraystretch}{1.15}
\resizebox{\columnwidth}{!}{
\begin{tabular}{cc | cc | cc}
\toprule
\multirow{2}{*}{\textbf{SFT}} & \multirow{2}{*}{\textbf{Retrieval}} & \multicolumn{2}{c|}{\textbf{MER2023}} & \multicolumn{2}{c}{\textbf{MER2024}} \\
\cmidrule{3-6}
& & \textbf{HIT} $\uparrow$ & \textbf{$\Delta$HIT} & \textbf{HIT} $\uparrow$ & \textbf{$\Delta$HIT} \\
\midrule
$\times$     & $\times$           & 20.24 & ---     & 18.32 & --- \\
$\checkmark$ & $\times$           & 24.12 & $+3.88$ & 22.45 & $+4.13$ \\
$\checkmark$ & Naive RAG          & 26.58 & $+6.34$ & 24.71 & $+6.39$ \\
\rowcolor{mygray} \textbf{$\checkmark$} & \textbf{AffectAgent} & \textbf{28.80} & \textbf{+8.56} & \textbf{27.53} & \textbf{+9.21} \\
\bottomrule
\end{tabular}
}
\vspace{-1em}
\end{table}
\begin{table}[t]
\centering
\caption{Ablation on Multi-Agent coordination and Evidence Structures based on Video-LLaMA2. All variants retain primary evidence. $\Delta$HIT indicates the performance drop compared to the full framework.}
\vspace{-1em}
\label{tab:agent_evidence}
\renewcommand{\arraystretch}{1.15} 
\resizebox{\columnwidth}{!}{
\begin{tabular}{cccc | cc | cc}
\toprule
\multirow{2}{*}{\textbf{Planner}} & \multirow{2}{*}{\textbf{Filter}} & \multirow{2}{*}{\textbf{Confuse Evid.}} & \multirow{2}{*}{\textbf{Counter Evid.}} & \multicolumn{2}{c|}{\textbf{MER2023}} & \multicolumn{2}{c}{\textbf{MER2024}} \\
\cmidrule{5-8}
& & & & \textbf{HIT} $\uparrow$ & \textbf{$\Delta$HIT} & \textbf{HIT} $\uparrow$ & \textbf{$\Delta$HIT} \\
\midrule
$\times$     & $\checkmark$ & $\checkmark$ & $\checkmark$ & 27.15 & $-1.65$ & 25.86 & $-1.67$ \\
$\checkmark$ & $\times$     & $\checkmark$ & $\checkmark$ & 27.94 & $-0.86$ & 26.65 & $-0.88$ \\
$\times$     & $\times$     & $\checkmark$ & $\checkmark$ & 26.58 & $-2.22$ & 24.71 & $-2.82$ \\
\midrule
$\checkmark$ & $\checkmark$ & $\times$     & $\times$     & 27.42 & $-1.38$ & 26.11 & $-1.42$ \\
$\checkmark$ & $\checkmark$ & $\checkmark$ & $\times$     & 28.11 & $-0.69$ & 26.75 & $-0.78$ \\
\rowcolor{mygray} \textbf{$\checkmark$} & \textbf{$\checkmark$} & \textbf{$\checkmark$} & \textbf{$\checkmark$} & \textbf{28.80} & \textbf{---} & \textbf{27.53} & \textbf{---} \\
\bottomrule
\end{tabular}
}
\vspace{-2em}
\end{table}

\vspace{-1em}
\subsection{Comparison with RAG Methods}
Table~\ref{tab:rag_method_comparison} compares AffectAgent with existing RAG methods under the same full-modality setting. Overall, AffectAgent achieves the best results on all datasets and obtains the highest mean score of 76.78, consistently outperforming Rewrite-Retrieve-Read, BGM, and RAG-DDR. This indicates that directly applying general RAG strategies to multimodal emotion recognition is insufficient, since the task requires not only semantically relevant retrieval but also affect-aware evidence selection and reasoning. The consistent gains across Basic Emotion Recognition, Sentiment Analysis, and Fine-grained Emotion Recognition further verify that AffectAgent can more effectively exploit external evidence for multimodal emotion understanding.
\vspace{-1em}
\subsection{Ablation Study}
Table~\ref{tab:main_results} shows the stepwise gains of our framework on MER2023 and MER2024, which with Video-LLaMA2 as the backbone. SFT already improves the zero-shot baseline by +3.88 and +4.13 HIT, and naive retrieval further boosts performance to +6.34 and +6.39, confirming the value of external evidence. Our full AffectAgent achieves the best results on both datasets, reaching 28.80 and 27.53 HIT, which means +8.56 and +9.21 over zero-shot. These results indicate that the gain comes not only from retrieval itself, but also from the proposed affect-oriented retrieval and reasoning framework.

Table~\ref{tab:agent_evidence} further examines the roles of multi-agent coordination and evidence structures, where all variants retain primary evidence. Removing either the Planner or the Filter degrades performance, with a larger drop from removing the Planner (-1.65/-1.67) than the Filter (-0.86/-0.88), while removing both causes the largest loss and reduces the model exactly to the naive retrieval setting in Table~\ref{tab:main_results}. Likewise, removing confusion or counter evidence also hurts performance, with confusion evidence contributing more than counter evidence. This shows that primary evidence alone is insufficient for robust emotion reasoning, and the full model performs best only when both agents and all evidence structures are jointly enabled. More ablation experiments can be found in the supplementary material.

\begin{figure}[thbp]
    \centering
    \includegraphics[width=\linewidth]{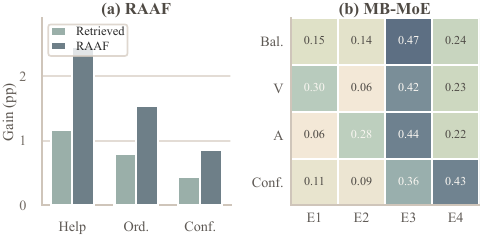}
    \caption{Visualization of RAAF and MB-MoE in AffectAgent on sampled MER2023 data with AffectGPT as the backbone. (a) Performance gain (pp) under different retrieval conditions. (b) Average expert routing weights under different input states.}
    \Description{Two plots summarize the fusion modules. The first compares performance gains under different retrieval conditions, and the second compares average MB-MoE expert-routing weights for different modality input states.}
    \label{fig:visualization_experiment}
    \vspace{-2.2em}
\end{figure}
\subsection{Visualization Experiments}
To examine RAAF and MB-MoE, we visualize their behavior on a randomly sampled subset of the MER2023 evaluation set using AffectGPT as the MLLM. The left plot in Fig.~\ref{fig:visualization_experiment} compares performance gains in percentage points from direct use of retrieved perceptual evidence and from RAAF under helpful, ordinary, and conflicting retrieval. The right plot shows the average routing weights of four experts under balanced, video-led, audio-led, and conflicting inputs.

RAAF produces larger gains than direct evidence use in all three conditions. The improvement is largest under helpful retrieval and remains clear under ordinary and conflicting retrieval. This indicates that RAAF uses complementary evidence without uniformly amplifying all retrieved information. The routing weights also vary with the input state. E3 remains dominant, while E1, E2, and E4 respond more strongly to video-led, audio-led, and conflicting inputs, respectively. These patterns show that the two modules adapt to the quality and modality balance of the input.

\subsection{Case Study}

Figure~\ref{fig:case_study} shows both the robustness and limitations of AffectAgent. In the left case, the generated queries align with the underlying emotional cues, allowing the retriever to return relevant evidence and produce a correct prediction. In the middle case, some query drift occurs during reasoning, but the retrieved evidence retains the dominant emotional signal and the model recovers the correct result. In the right case, the drift introduces misleading evidence that gradually affects later evidence selection and reasoning and leads to an incorrect prediction. These cases show that AffectAgent tolerates moderate errors in intermediate decisions but remains sensitive when the evidence is severely distorted or misaligned with the true emotional state.

\vspace{-1em}
\section{Conclusion}
We propose AffectAgent, an affect-oriented multi-agent retrieval-augmented framework for multimodal emotion recognition. It coordinates query planning, evidence filtering, and emotion generation for reliable fine-grained reasoning. MB-MoE balances modality contributions, while RAAF supplements missing information with retrieved evidence. Together, they improve cross-modal fusion and robustness under incomplete modalities, providing an effective solution for accurate multimodal emotion recognition.

\begin{acks}
This work was supported by the National Natural Science Foundation of China (Grant No.~62576076), the CCF-Tencent Rhino-Bird Open Research Fund, the Guangdong Research Team for Communication and Sensing Integrated with Intelligent Computing (Project No.~2024KCXTD047), the Open Research Fund from Guangdong Laboratory of Artificial Intelligence and Digital Economy (SZ) (Grant No.~GML-KF-26-17), the Guangdong Basic and Applied Basic Research Foundation (Grant Nos.~2026A1515010184, 2023A1515140037, and 2025B1515120017), and the Guangdong Provincial Key Laboratory (Grant No.~2023B1212060076). The computational resources were supported by the SongShan Lake HPC Center (SSL-HPC) at Great Bay University.
\end{acks}

\bibliographystyle{ACM-Reference-Format}
\bibliography{sample-base.bib}


\end{document}